\pdfoutput=1

\documentclass[11pt]{article}

\usepackage[final]{acl}

\usepackage{times}
\usepackage{latexsym}

\usepackage[T1]{fontenc}

\usepackage[utf8]{inputenc}

\usepackage{microtype}

\usepackage{inconsolata}

\usepackage{graphicx}

\usepackage{subcaption} 
\usepackage{booktabs} 
\usepackage{array}    
\usepackage{multirow}
\usepackage{amsmath}
\usepackage{hyperref}
\title{Recurrent Context Compression: Efficiently Expanding the Context Window of LLM}

\author{
	\textbf{Chensen Huang\textsuperscript{1}},
	\textbf{Guibo Zhu\textsuperscript{2,3}},
	\textbf{Xuepeng Wang\textsuperscript{2,3}},
	\\
	\textbf{Yifei Luo\textsuperscript{1}},
	\textbf{Guojing Ge\textsuperscript{2,3}},
	\textbf{Haoran Chen\textsuperscript{1,2}},
	\textbf{Dong Yi\textsuperscript{2,3}},
	\textbf{Jinqiao Wang\textsuperscript{2,3}}
	\\
	\\
	\textsuperscript{1}University of Chinese Academy of Sciences,
	\\
	\textsuperscript{2}Institute of Automation, Chinese of Academy,
	\\
	\textsuperscript{3}Wuhan AI Research
	\\
	\\
	\small{
		\textbf{Correspondence:} \href{mailto:ken@yuniversity.edu}{huangchensen2022@ia.ac.cn}
	}
}

\begin{document}
\maketitle
\begin{abstract}
To extend the context length of Transformer-based large language models (LLMs) and improve comprehension capabilities, we often face limitations due to computational resources and bounded memory storage capacity. This work introduces a method called Recurrent Context Compression (RCC), designed to efficiently expand the context window length of LLMs within constrained storage space. We also investigate the issue of poor model responses when both instructions and context are compressed in downstream tasks, and propose an instruction reconstruction method to mitigate this problem. We validated the effectiveness of our approach on multiple tasks, achieving a compression rate of up to 32x on text reconstruction tasks with a BLEU4 score close to 0.95, and nearly 100\% accuracy on a passkey retrieval task with a sequence length of 1M. Finally, our method demonstrated competitive performance in long-text question-answering tasks compared to non-compressed methods, while significantly saving storage resources in long-text inference tasks. Our code, models, and demo are available at \url{https://github.com/WUHU-G/RCC\_Transformer}

\end{abstract}

\begin{figure*}[h]
	\centering
	\includegraphics[width=1\linewidth,keepaspectratio]{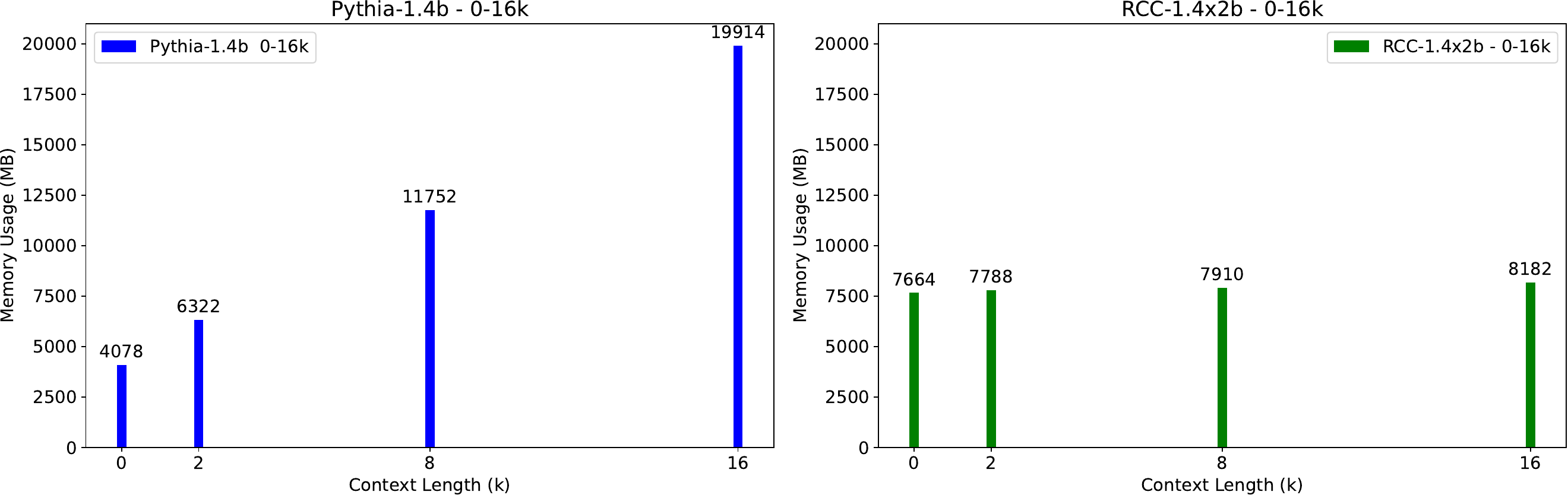}
	\caption{GPU memory Consumption of Different Models with Increasing Length. Left: Pythia-1.4b, Right: RCC model using Pythia-1.4b for both encoder and decoder. Both models utilize FlashAttention-2 \cite{dao2023flashattention2}. A more detailed analysis of GPU memory consumption can be found in Appendix \ref{ap5}.} 
	\label{fig8}
\end{figure*}

\section{Introduction}

With the rapid advancement of natural language processing technologies, Transformer-based large language models (LLMs) have become a key driving force in this field. However, when handling long text inputs, LLMs often encounter limitations in context window length. These limitations stem from several inherent factors in the model architecture and training methods. Firstly, during the inference phase, models are constrained by the pretraining text length, leading to a significant decline in quality when the generated sequence exceeds the pretrained context window. Secondly, the design of the Transformer architecture requires storing information from the entire input sequence, which results in a substantial memory footprint due to the KV-Cache during inference.

To address these issues, related research works \cite{c31,c23,c27,c25,c26,c20,c29,c24} have optimized training methods, model structures, and KV-Cache optimization, thereby extending the context window of LLMs. Among these, context compression techniques \cite{c25,c5,c6,c4,c3,c7,c9,c34,c35} are considered promising because they can compress context or prompts into shorter forms while maintaining good performance, thus enabling the inference of longer context windows within limited resources. Figure \ref{fig8} compares the memory resource consumption of our method with non-compression methods. Additionally, most text compression-based works can be integrated and combined with other context window extension techniques to enhance performance.

However, existing context compression methods face three major challenges in long-text language modeling. Firstly, the efficiency of compression has certain limitations. For example, ICAE with 14B parameters \cite{c7} experiences a significant performance drop beyond an 8x compression rate. Secondly, most context compression research focuses on shorter sequences rather than long texts. We found that language models trained for context compression on short sequences perform poorly when directly extended to long sequences, necessitating new methods to improve long-text compression performance. Lastly, in practical applications, we observed that context-compressed language models face the issue of context-instruction confusion in downstream tasks. When both context and instructions are compressed simultaneously, the model often struggles to follow instructions correctly, resulting in poor controllability. This issue has not been emphasized or addressed in previous studies, which typically compress only context or instruction texts individually.

To address the aforementioned issues, this paper makes the following contributions:

Firstly, we propose a context compression model structure based on an autoencoder, which we call the Recurrent Context Compression (RCC) architecture. RCC significantly reduces information loss during the compression process, greatly enhancing compression efficiency. In experiments, we achieved nearly 95\% BLEU-4 scores with a 32x compression rate on text reconstruction tasks. Although the encoder occupies some memory resources, the memory required by traditional language models exceeds that of the context compression model when the sequence length surpasses a certain threshold, as shown in Figure \ref{fig8}.

Secondly, we propose a new training method to adapt long-text context compression language models. We introduce a recurrent compression mechanism to overcome the context window limitations of the encoder, allowing the model to compress texts beyond the encoder window length. When training long-text context compression language models, the length of the context sequence input to the encoder is proportional to the required computational resources, limiting the extension of context length during training. Therefore, we propose a simple yet effective solution: initially, conducting full-parameter training on shorter sequences. Subsequently, we freeze the encoder on the saved model weights and continue training on longer sequences, enabling the extension of training context length under constrained computational resources.The detailed information can be found in Section \ref{sec1}.

Lastly, in downstream text generation tasks, we found that when both context and instruction texts are compressed, the model struggles to follow instructions properly, leading to a decline in response quality. To mitigate this issue, we leverage the text reconstruction capability of the context compression language model, allowing the decoder to reconstruct the instruction content from the compressed vectors and continue generating responses based on the instructions. This significantly improves the output quality when both context and instructions are compressed, achieving results close to those obtained by inputting instructions directly into the encoder.

\section{Related Work}
\textbf{Context Compression:} Early approaches to context compression aimed to derive sentence representation vectors for tasks such as document retrieval. Transformer-based autoencoder architectures like TSDAE \cite{c1} and Nugget \cite{c2} are relevant to our work. In TSDAE, noise such as word deletion or swapping is added to input sentences to train sentence embedding vectors. The encoder compresses the corrupted sentence into a fixed-size vector, which the decoder then reconstructs into the original input text. However, such approaches cannot be directly applied to text generation tasks. Gist\cite{c3} leverages Transformer-based large language models (LLMs) as autoencoders. During training, a clever masking matrix compresses prompts into a few Gist tokens, which can still prompt the language model for responses. Similar prompt compression work was proposed by \cite{c4}. However, these tasks only compress prompts.

Several works \cite{c5,c6,c7,c34,c35,c8,c9} have focused on context compression. ICAE\cite{c7} is similar to our work but suffers from lower compression efficiency and lacks extensive research on longer sequences. Additionally, \cite{c8} employed non-vector-based context compression by using smaller language models to input long texts and generate more compact short texts. Selective context, proposed by \cite{c35}, identifies and prunes redundancy in the input context to enhance LLM inference efficiency, making inputs more compact. Recently, \cite{c9} proposed a similar work combining two attention mechanisms with context compression functionality, showing promising results. However, this new attention mechanism cannot be directly applied to pre-trained open-source LLMs and faces attention optimization challenges in practical applications. None of the studies have thoroughly investigated the problem of instructional confusion that arises when both instructions and contextual information are subjected to compression. Our work introduces a new solution to mitigate this issue. 

Additionally, our work is inspired by language models with recurrent structures \cite{c31,c10,c11,c12}. These models compress historical context within a certain range into the hidden state of a single time step, enabling the current token to access information from the previous step for inference. They demonstrate strong competitiveness with Transformer models, indicating that compressing token information over a certain length can achieve lossless inference.

\textbf{Long Context LLM:} LLMs typically fix the context window length during training, such as the Pythia \cite{c13}, LLaMA \cite{c14,c15}, and Mistral \cite{c16} series. Consequently, researchers have explored various methods to extend the context window length of pre-trained language models. These methods\cite{c17,c18,c19,c20,c21}, which have achieved notable results based on existing pre-trained models. Our approach combines these methods, allowing us to apply them to the encoder or decoder to achieve more extended compression effects.

\begin{figure*}[h]
	\centering
	\includegraphics[width=1\linewidth,keepaspectratio]{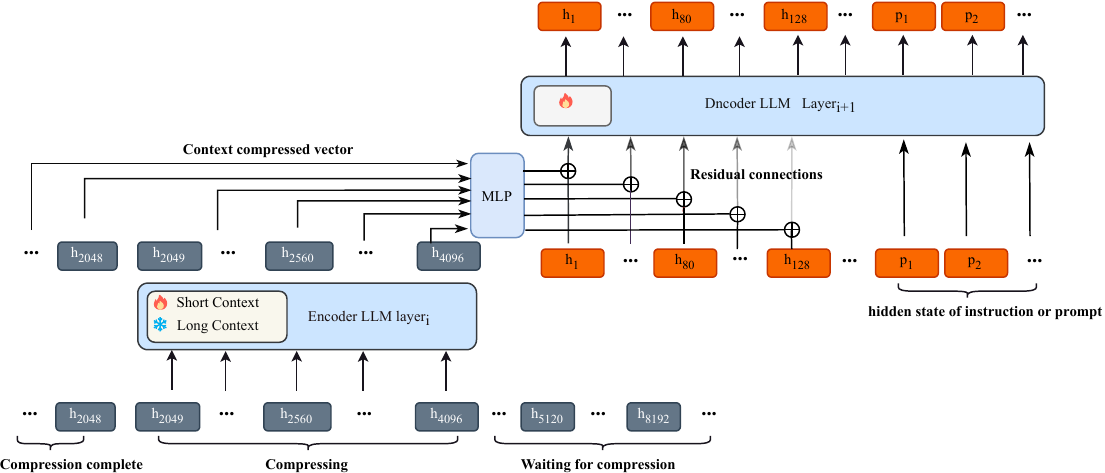}
	\caption{The structure of the encoder and decoder in RCC layer i. The maximum context window of the encoder is 2048. The encoder has a compression rate of 32, and we use the vectors at positions that are multiples of 32 in the output as the compressed vectors. Each segment will generate a compressed vector of length 64. When the sequence length exceeds 2048, the encoder performs cyclic segmentation and compression. The compressed vectors produced between segments in the encoder are independent, while those generated within a segment are correlated. The decoder's input is the residual connection between the input vector from the previous layer and the compressed vector after linear mapping. All compressed vectors will interact within the decoder.}
	\label{fig1}
\end{figure*}		

\section{Design of RCC}
\subsection{Method Overview}
Our encoder design is inspired by the Mamba-based LLM \cite{c10}. Mamba is essentially a state space model, similar to an RNN. In Mamba, the current token only needs to access the state vector from the previous timestep to complete the current inference step. However, as the context length increases, the performance of Mamba deteriorates. This indicates that the state vector at each timestep in Mamba can store only a limited length of historical context information. Therefore, we propose a compromise: for long sequences, we can divide them into fixed-length short sequences and iteratively compress each short sequence into a state vector. We concatenate the state vectors of each short sequence as the historical state information during inference. This approach maximizes the retention of complete historical information while leveraging the model’s compression capabilities to save memory. In this paper, we use compression rate to reflect the maximum context length that a state vector at each timestep can store. Our experiments show that Transformers also have this capability because a Transformer can be viewed as a special state space model or RNN. Recent studies have shown that attention can be viewed as an RNN \cite{feng2024attention}.

\subsection{RCC Model Architecture}
As shown in Figure \ref{fig1}, the RCC model architecture is similar to ICAE\cite{c7}, consisting of an encoder and a decoder. Unlike the connection method in ICAE, where the final layer vector of the encoder is used as the input to the decoder, we take a different approach. We use the output information from each layer of the encoder. This information is then linearly mapped and input into the decoder. This method obtains more feature information, and the relevant ablation experiments can be found in Figure \ref{fig:img0}. The encoder can be a Transformer-based LLM or an RNN-based LLM, while the decoder is a Transformer-based LLM. The encoder is responsible for compressing the information, and the decoder reads the compressed information and performs inference. The decoder can fully learn the compressed information vector at any position using the attention mechanism. After proper training, the maximum context length that the RCC model can support is the encoder compression rate multiplied by the decoder context window length.

\subsubsection{RCC Encoder}
The primary task of the RCC encoder is to compress long sequences. The initialized encoder is a pretrained language model, which can be based on either Mamba or Transformer architectures. By setting a compression rate, we divide long sequences into fixed-length short sequences and iteratively feed these short sequences into the encoder. As illustrated in Figure \ref{fig1}, we locate the token at each position multiple of the compression rate within each short sequence. The output vectors of these tokens from each layer serve as compressed vectors, which are concatenated to form the final compressed vector for the entire sequence. Through this method, the RCC encoder accomplishes the compression modeling of the entire long context.

\subsubsection{RCC Decoder}
The decoder of the RCC is a Transformer-based language model responsible for the final text inference. Its inputs include the compressed vectors from the encoder and token embedding vectors related to the prompts. Each layer's compressed vector from the encoder passes through a linear layer before being input into the decoder. For the first layer's mapped vector, we concatenate it with the decoder's token embedding vector and then feed it into the first block of the decoder. Subsequently, the output part of each block is connected with the corresponding layer's compressed vector through residual connections. It is crucial to note that only the output vectors corresponding to the compressed information will have residual connections, while the other output parts remain unchanged. If the number of compressed vector layers does not match the number of decoder layers, we apply simple rule-based mappings, either by duplicating to increase the number of layers or averaging to reduce the number of layers.

\subsection{Model Training Tasks}
For the model training tasks, we require the model to possess both contextual memory and contextual reasoning abilities. Therefore, we selected text reconstruction and text continuation tasks. Traditional autoencoding text reconstruction tasks \cite{c7} are not suitable for text generation paradigms, so we replaced them with a random prompt text reconstruction task. Specifically, we randomly extract a short text segment from the encoder's input text as a prompt to the decoder, requiring the decoder to reconstruct the content following the prompt by leveraging the compressed information and the prompt. This task enhances the model's memory ability. Additionally, to strengthen the model's reasoning ability, we employed text continuation tasks. Relevant formulas can be found in Appendix \ref{ap1}.

\subsubsection{Long Text Training Methods}
\label{sec1}
We adopt a cyclic segmentation approach for computing long sequences. During model inference, we only need to store the compressed vectors. However, during training, we also need to store the gradient information for the entire long sequence, which significantly exceeds memory limits. We mitigate this issue using a simple yet effective two-stage training method. In the first stage, we perform full-parameter fine-tuning with a large number of shorter sequences, allowing the encoder to sufficiently learn how to compress the context into a single vector. Correspondingly, the decoder learns to infer or reconstruct from the compressed vectors. After the first stage, the encoder is capable of producing standardized compressed vectors. At this point, we input longer sequences and freeze the encoder, enabling the decoder to learn how to infer from more compressed vectors. This method does not require complex gradient optimization algorithms or substantial GPU memory resources\cite{c20,c6,c20} and can efficiently scale to longer sequences. We validated the effectiveness of this method on a 1M-length key retrieval task.

\subsubsection{Instruction Reconstruction Method}
To address the poor performance when both instructions and text are compressed simultaneously, we propose an instruction reconstruction method. During the fine-tuning phase, we input the instruction as part of the context to the model's encoder, with the instruction randomly placed at the beginning or end of the context. The decoder is then required to first reconstruct the instruction and subsequently answer the question based on the instruction.

\section{Experiments}

First, we conducted text compression rate experiments using the random prompt text reconstruction task, selecting ICAE \cite{c7} as the baseline model. Subsequently, we evaluated our method's performance on long text tasks, including the passkey context block retrieval task with 1M characters and the long document question-answering benchmark in Longbench \cite{c28}.For model architecture, we used pythia-1.4b \cite{c13} as the encoder and tested mamba-1.4b \cite{c10}, both supporting a 2048 context window. The decoder was pythia-1.4b. We randomly sampled about 5 billion tokens from the pile \cite{c33} dataset as the training set, concatenating these tokens into a continuous ultra-long one-dimensional array. The learning rate was set to 1e-4. Training was stopped if the model failed to converge after one epoch or converged prematurely within an epoch, typically completing within 30 hours on a server with 8 A800 GPUs.

\begin{figure*}[htbp]
	\centering
	\begin{subfigure}{0.48\textwidth} 
		\centering
		\includegraphics[width=\linewidth,keepaspectratio]{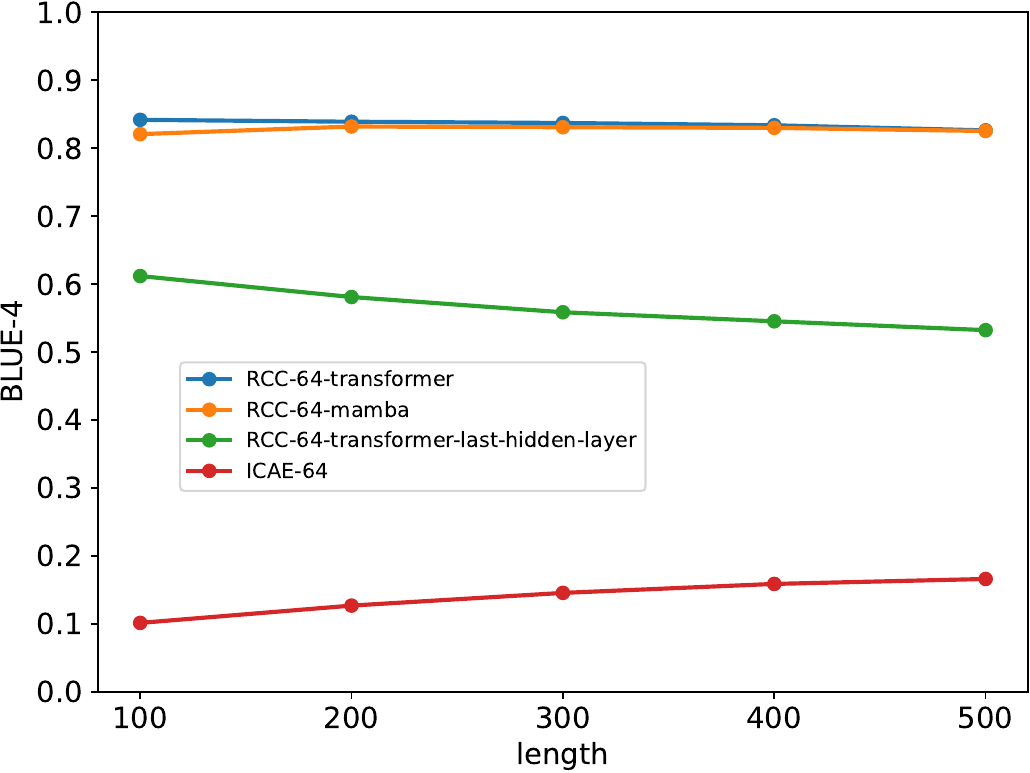}
		\caption{Text Reconstruction Scores of Different Models.}
		\label{fig:img0}
	\end{subfigure}\hfill
	\begin{subfigure}{0.48\textwidth}
		\centering
		\includegraphics[width=\linewidth,keepaspectratio]{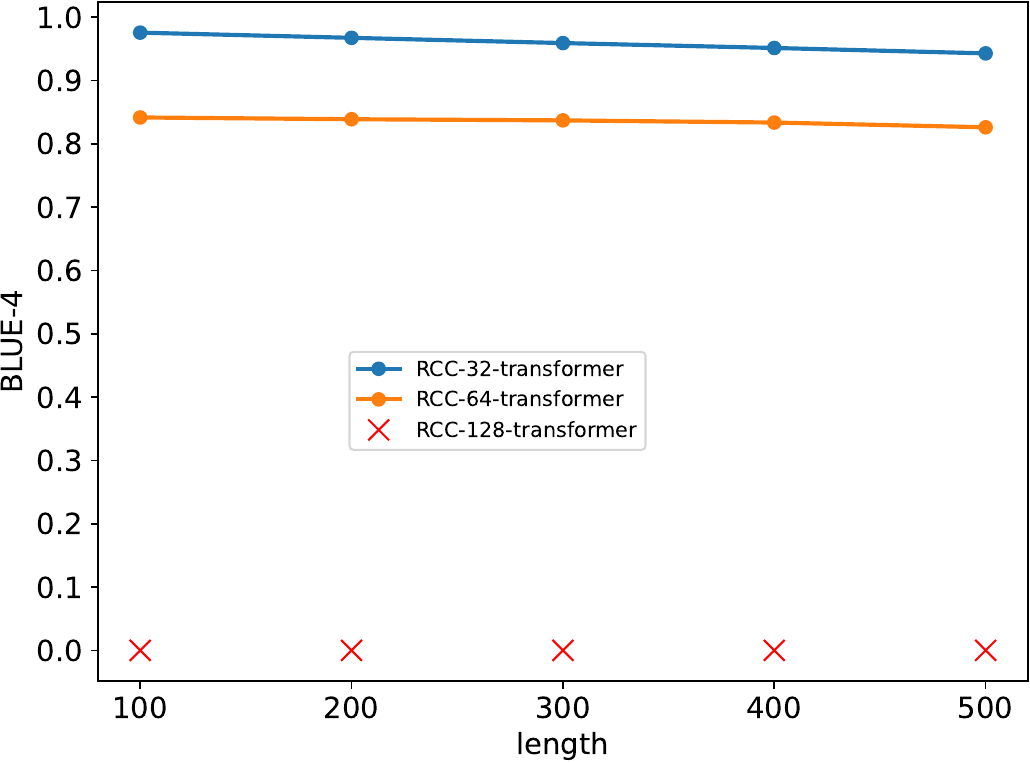}
		\caption{Scores at Different Compression Rates}
		\label{fig:img1}
	\end{subfigure}
	\caption{Text reconstruction score}
	\label{both_images}
\end{figure*}

\subsection{Text Reconstruction}

In the initial training phase, we used a combination of random prompt text reconstruction tasks and Text Continuation tasks, with a ratio of 9:1. This was followed by fine-tuning with the random prompt text reconstruction task. The input sequence length for the encoder was set to 2048 tokens, while the uncompressed part of the decoder had an input length of 512 tokens. For the random prompt text reconstruction task, the uncompressed part of the decoder's input was a subset of the encoder's input sequence, including prompts and the text to be reconstructed. The prompts formed the initial part of this subsequence and were excluded from the loss calculation; only the loss of the reconstructed text following the prompts was calculated. In the Text Continuation task, the decoder's uncompressed input sequence was the continuation of the encoder's input sequence. We assessed the model's compression performance using the BLEU-4 score, comparing the reconstructed text to the actual text. To achieve this, we created 100 encoder input samples, each with a token length of 2048. To ensure fairness in the scores, we selected 5 text segments as prompts for the decoder at every 300-token interval from the sample. Each decoder then calculated the reconstruction score for prompts at 5 different positions, and we averaged these scores. The prompt text length was about 10 tokens, and the reconstruction text length was about 500 tokens. Reconstruction examples can be found in Appendix \ref{ap3}.

As shown in Figure\ref{fig:img0}, we compared different models under a 64× compression rate. Both RCC-64-mamba and RCC-64-transformer achieved a BLEU-4 score close to 0.82, but mamba’s training time was nearly 1.5 times that of transformer. RCC-64-transformer-last-hidden-layer, which uses only the encoder's last layer compression vectors, achieved a BLEU-4 score of approximately 0.6. This approach, common in traditional autoencoder models \cite{c1,c2,c7}, retains less textual information compared to using compression vectors from all layers. Additionally, ICAE performed poorly under a 64× compression rate, with a BLEU-4 score of about 0.1, confirming our method's effectiveness in preserving text information.

As shown in Figure \ref{fig:img1}, we tested reconstruction performance under different compression rates. At a 32× compression rate, the BLEU-4 score reached 0.95. At a 64× compression rate, the score dropped to between 0.8 and 0.85. At a 128× compression rate, we encountered convergence issues, preventing BLEU-4 score computation. This indicates that higher compression rates increase the difficulty of text reconstruction.

\begin{table*}[htbp]
	\centering
	\begin{tabular}{l *{5}{p{1.5cm}}}
		\toprule
		\textbf{Model} & \textbf{32K} & \textbf{128K} & \textbf{256K} & \textbf{512K} & \textbf{1M} \\
		\midrule
		RCC-Mamba-FT-8k & 98/100/97 & 96/98/94 & 95/93/89 & 94/95/96 & 94/96/96A \\
		RCC-Transformer-FT-8k & 97/95/96 & 96/97/96 & 92/96/96 & 92/89/95 & 97/96/96 \\
		RCC-Mamba-FT-32k & 100/100/100 & 100/100/100 & 100/100/100 & 100/100/100 & 100/99/100 \\
		\textbf{RCC-Transformer-FT-32k} & 99/100/100 & 100/100/100 & 100/100/100 & 98/100/100 & 100/100/100 \\
		Infini-Transformer-FT & 100/100/100 & 100/100/100 & 100/100/100 & 97/99/100 & 96/94/100 \\
		\bottomrule
	\end{tabular}
	\caption{The performance of the different models on the passkey retrieval tasks ranging from 32k to 1M sequence lengths, RCC-512-FT-8k denotes that the RCC model is trained with full parameters on a fine-tuning dataset with a length of 8k. RCC-512-FT-64K is trained on a fine-tuning dataset with a length of 64K based on RCC-512-FT-8k, while in this case, we freeze the encoder.}
	\label{tab2:performance}
\end{table*}

\subsection{Passkey Retrieval Task}
We utilize the passkey retrieval task \cite{c29} to validate the effectiveness of the two-stage training method mentioned in section \ref{sec1}. Additionally, we observe that our method exhibits certain length extrapolation capabilities, enabling it to handle compressed vector lengths during inference that far exceed those seen during training. This indicates that the compressed vectors generated by our method can be reliably recognized by the encoder, with minimal influence from positional encoding. Passkey retrieval task involves embedding a random number into a long sequence composed of repeated fixed short phrases, with the overall text length controlled by adjusting the number of repetitions of these short phrases. The task requires the model to accurately retrieve the hidden number from these long sequences. Detailed construction methods for passkey retrieval task samples are provided in Appendix \ref{ap2}. In this task, we employed a compression rate of 512x. Although this compression rate might not be effective for reconstruction tasks, experiments show that the fine-tuned RCC model performs well in the passkey retrieval task. The model was first pre-trained on a dataset containing only the random prompt text reconstruction task, with an encoder input length set to 8k and a non-compressed decoder input length of 512. After pre-training, we constructed nearly 30,000 passkey retrieval task samples with context lengths of 8k and 32k, respectively. These samples formed the fine-tuning dataset. We conducted a two-stage fine-tuning process. In the first stage, we fine-tuned the entire parameter set using the 8k context length samples. After completing the first stage, we proceeded to the second stage with the 32k context length samples. During this stage, we froze the encoder parameters to accommodate the constraints of limited available memory.

From Table\ref{tab2:performance}, we can observe that even with the encoder using only an 8k context window, the model achieves almost 90\% accuracy in passkey retrieval tasks up to 1000k, demonstrating the strong length extrapolation capabilities of our model. After the second stage of fine-tuning with sequences up to 32k, the model achieves nearly 100\% performance on passkey retrieval tasks up to 1000k, proving the effectiveness of our two-stage training method, even with the encoder parameters frozen at this stage. Table\ref{tab2:performance} also shows that our method is highly competitive compared to recent similar work like Infini-attention \cite{c9}. Unlike Infini-attention, our method can be fine-tuned on existing open-source LLMs with a small amount of data, without requiring the reconstruction of the LLM model and pre-training with hundreds of billions of tokens.

\begin{table*}[ht]
	\centering
	
	\begin{tabular}{cccccccccc}
		\toprule
		\multirow{1}{*}{Method} 
		& 0-2k & 2-4k & 4-8k & 8k+  & average  \\
		\midrule
		
		Pythia-SFT  & \textbf{30.54} & - & - & - & - \\
		Pythia-No-SFT  & 4.41 & - & - & - & - \\
		RCC-Ins-Reconstruction  & \textbf{28.12} & 23.37 & 21.24 & 17.72 & 22.61 \\
		RCC-Ins-Human  & 25.36 & \textbf{25.15} & \textbf{23.63} & \textbf{20.48} & \textbf{23.15} \\
		RCC-Ins-Compress  & 18.77 &  21.36 & 20.02 & 18.14 & 19.61\\

		\bottomrule
	\end{tabular}
	\caption{Scores of different models on the task of Document QA. }
	\label{tab:performance}
\end{table*}
\subsection{Long-Text Benchmark Evaluation}

\subsubsection{Evaluation Dataset} 
LongBench \cite{c28} is a benchmark designed to evaluate the capabilities of large language models in understanding long contexts. To assess our model's performance on texts of different lengths, we selected the LongBench-E set for evaluation because it evenly covers test samples of various length ranges, allowing us to analyze the impact of length variation on performance. Due to limitations in the fine-tuning dataset, our work focuses on using the single-document QA and multi-document QA tasks for evaluation. These two document QA tasks consist of four subtasks, with each task containing between 150 to 300 samples. Detailed information on the evaluation dataset can be found in Appendix \ref{ap4}.



\subsubsection{Instruction Fine-Tuning}
First, we conducted pretraining on the random prompt text reconstruction task and the text continuation task with a ratio of 1:9. Similar to the two-stage training method, the first stage involves training with full parameters on texts with a length of 2k. In the second stage, the encoder is frozen, and training is conducted on texts with a length of 16k. For question answering instruction fine-tuning, we used the Prompt-with-Context (PwC) \cite{c7} and hotpotQA \cite{hotpotqa} datasets. These datasets include context with instructions and outputs, teaching the model to use context for answering questions rather than relying solely on internal knowledge. We concatenated the context and instructions as the encoder's input, while the instructions and output results formed the decoder's uncompressed input. We repeated the instructions twice to train the encoder to reconstruct instructions, enhancing mixed instruction and context text effectiveness during inference. The PwC dataset has 240k samples, and hotpotQA has 90k samples. Additionally, to improve instruction reconstruction and maintain the decoder's instruction-following capability, we randomly selected 50k instruction samples from the orcal dataset \cite{orca}. These samples lack explicit context fields and typically mix instructions with context. We input the instructions as context to the encoder and as instructions and outputs to the decoder. During fine-tuning, the encoder's input context length was set to 2048 tokens, the uncompressed part of the decoder's input was set to 512 tokens, and the compression rate remained 32.


\subsubsection{Evaluation}

We fine-tuned Pythia-1.4b with instruction pairs constructed from PwC, hotpotQA, and some ORCA data to ensure it follows instructions. Due to limitations in the fine-tuning dataset, we only selected document QA tasks. We evaluated the fine-tuned Pythia-1.4b and our model using LongBench's\cite{c28} automated evaluation tools, covering two document QA tasks as shown in Appendix \ref{ap4}. As shown in Table \ref{tab:performance}, the fine-tuned Pythia-1.4b significantly improved in following instructions. Notably, Pythia-1.4b supports a maximum sequence length of 2048 tokens, so we only used samples under 2k tokens for its evaluation. Our method supports LongBench's maximum input length of 15k tokens within the effective window length of the decoder. We further evaluated the following types for our method:


\textbf{RCC-Ins-Reconstruction}, which reconstructs instructions from compressed vectors and responds using instruction reconstruction techniques (Table 2), scored 28.12 at a length of 2k. This score is competitive with Pythia-sft, demonstrating that RCC can maintain high-quality inference even with a compression ratio of up to 32x. This method's average score surpasses that of RCC-Ins-Compress, which compresses both instructions and context, verifying the effectiveness of instruction reconstruction. Due to the fine-tuning dataset being limited to 2k tokens, RCC-Ins-Reconstruction performs poorly in instruction reconstruction when handling longer samples.

\textbf{RCC-Ins-Human directly} inputs real instruction texts into the decoder (Table 2). Compared to the performance fluctuations of RCC-Ins-Reconstruction with increasing sample length, RCC-Ins-Human exhibits more stable performance, especially maintaining efficient inference at lengths beyond 8k. We attribute this to the decline in instruction reconstruction quality in RCC-Ins-Reconstruction for long texts, whereas RCC-Ins-Human employs fixed instructions, unaffected by length.

\textbf{RCC-Ins-Compress compresses} both context and instructions simultaneously (Table 2). The encoder receives concatenated texts, and the decoder is only prompted with brief information, such as "\\n Answer:". This strategy's limited capability to recognize instructions and context results in an average score as low as 19.61, particularly underperforming compared to RCC-Ins-Human and RCC-Ins-Reconstruction in samples under 8k. However, for ultra-long samples (8k+), its performance converges with RCC-Ins-Reconstruction, likely due to the latter's deficiencies in instruction reconstruction at extreme lengths. Specific model generation results can be found in Appendix \ref{ap4}.

In Figure \ref{fig8}, we observe that when RCC processes text up to 16k tokens, the GPU memory usage only increases by approximately 0.5 GB. In contrast, the original Pythia-1.4b experiences a 2 GB increase in memory usage for 2k token text. When processing 16k token text, Pythia-1.4b’s total memory usage will be twice that of our method. Since we use a compression rate of 32x, as text length increases, RCC can save up to nearly 32x in storage space. Although the parameter count of RCC's encoder and decoder is twice that of Pythia, the impact of the model's parameter count on storage space significantly diminishes with increased text length.

\section{Conclusion}
Utilizing compression techniques to mitigate challenges in long-text training and inference has proven to be a highly promising strategy. Our work quantitatively analyzes the impact of different context lengths on compression performance, while also achieving higher compression rates than previous methods, thereby significantly enhancing the ability of large language models (LLMs) to handle long texts. The RCC method demonstrated outstanding performance across multiple test tasks, particularly in context compression reconstruction, long-document question answering, and key-context block retrieval tasks with sequences up to 1 million tokens. Additionally, we analyzed the issues arising from simultaneous compression of context and instructions and introduced an instruction reconstruction method that effectively alleviated these problems. Furthermore, to address the substantial resource consumption of long-text training, we proposed a staged training strategy that further improved the efficiency of the model in handling long-text training.

\section{Limitations}


The RCC method has significantly advanced text compression efficiency and long-document question answering, but it has limitations. For example, RCC risks errors during instruction reconstruction, and if instructions are too long, the decoder may struggle to reconstruct them within the limited window. In future research, we plan to adopt a hybrid training approach: using instruction compression for long instructions and instruction reconstruction for short ones to achieve results comparable to manually input instructions.Additionally, the lack of long-text instruction fine-tuning data has caused performance bottlenecks for RCC. Our experiments show the critical impact of training data on the model’s performance. The effectiveness of language models fine-tuned with instructions depends largely on the quality and coverage of those instructions. These issues provide clear directions for our future research.

\bibliography{acl_latex_archav2.bib}

\appendix

\begin{figure*}[h]
	\centering
	\includegraphics[width=1\linewidth,keepaspectratio]{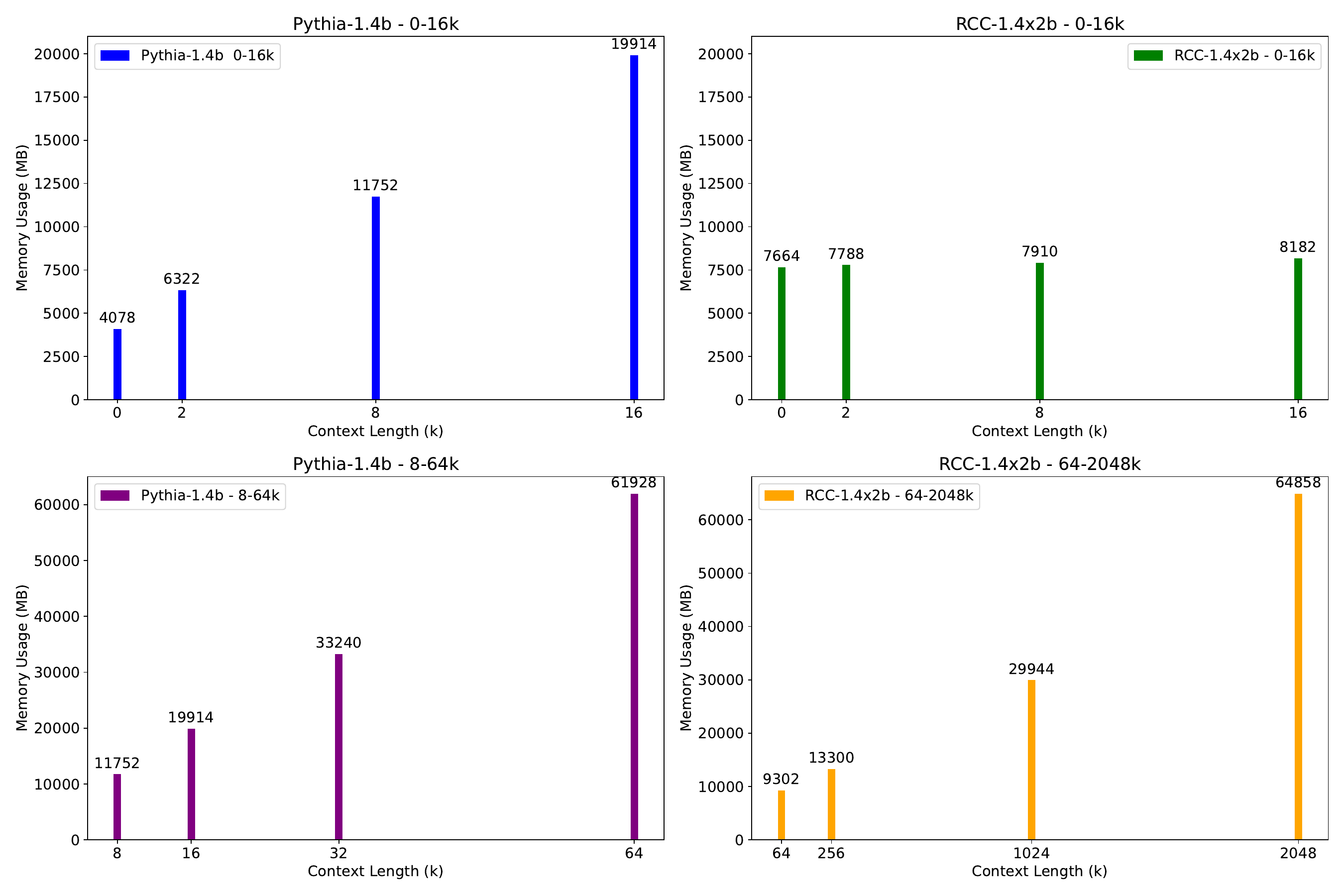}
	\caption{
		When the GPU memory approaches 60GB, the memory occupation of different models. Left: Pythia-1.4b, Right: RCC model using Pythia-1.4b for both encoder and decoder. Both models utilize FlashAttention-2 \cite{dao2023flashattention2}.} 
	\label{fig9}
\end{figure*}

\section{Effects of Text Reconstruction}	
\label{ap3}

The example of our method’s reconstruction effect at 32x compression rate is shown below. As the table \ref{table_co} indicates, our method has almost completely reconstructed the context.

\section{GPU Memory Consumption Analysis}	
\label{ap5}

We ran the model on an A800 GPU using HuggingFace's Transformers library \cite{wolf-etal-2020-transformers} and tested the GPU memory consumption of different models. As shown in Figure \ref{fig9}, the GPU memory usage of Pythia-1.4b increases rapidly with the length of the input context. When the context window reaches 64k, the model's GPU memory usage exceeds 60GB. RCC-1.4 x 2b, where both the encoder and decoder are Pythia-1.4b models with a compression rate of 32, shows that when its GPU memory usage exceeds 60GB, it processes a context length close to 2048k tokens. This is 30 times the length Pythia-1.4b can handle, nearly matching the compression rate.

\section{Random Prompt Text Reconstruction Tasks}
\label{ap1}

The random prompt text reconstruction tasks involves an original text sequence $({w_{1}}, \ldots, w_{n})$, where the encoder compresses the entire original text and produces a compressed hidden vector $({H})$. The decoder then needs to reconstruct the text that follows the random prompt word in the original text based on the encoder's input vector and the random prompt word from a segment of the original text. The prompt word is a substring of the original text $({w_{i}}, \ldots, w_{j})$, denoted as ${p}$, and the target sentence to be reconstructed, which follows the prompt word, is $({w_{j}}, \ldots, w_{x})$, denoted as ${c}$.

\begin{equation*}
	\mathcal{L}_{\mathrm{RAE}} = \max _{h, \ldots, p} P\left(\boldsymbol{c} \mid h, \ldots, p ; \Theta_{L L M}\right)
\end{equation*}

In the text continuation task, the prompt is no longer a substring of the encoder's input text but is the immediately following segment of text, and the target sentence is still the text that comes right after the prompt. The formula for the text continuation task is the same as that for the random prompt text reconstruction task.

\section{Format of Passkey Retrieval}
\label{ap2}
We follow the text format for passkey retrieval from existing works \cite{c17,c29}. The format of the document is as follows:

\texttt{There is an important info hidden inside a lot of irrelevant text.}
	\texttt{Find it and memorize them. I will quiz you about the important information there.}
	
	\texttt{The grass is green. The sky is blue. The sun is yellow. Here we go. There and back again. (repeat M times)}
	
	\texttt{The pass key is \textbf{56994}. Remember it. \textbf{56994} is the pass key.}
	\texttt{The grass is green. The sky is blue. The sun is yellow. Here we go. There and back again.}
	\texttt{(repeat N times)}
	
	\texttt{What is the pass key? The pass key is}

\begin{table*}[!htbp]
	\centering
	\begin{tabular}{ p{0.45\linewidth} p{0.45\linewidth} }
		\textbf{Our Result on RCC-32-Transformer} & \textbf{Standard Result} \\
		\hline
		{The Access nodes and storage daemons make up a data plane, while the core provides its control plane.  Also: How IBM Watson is revolutionizing 10 industries TechRepublic  So, what does all mean for customers? It\'s multi-cloud storage management, which enables allows you to manage, deploy, and migrate data storage across private and major public clouds. This includes Alibaba, AWS, Azure, and Google Cloud.  It\'s easy to see why Red Hat values this. It gives their customers a way to manage storage without sweating the details across multiple platforms.  As Ranga Rangachari, Red Hat\'s vice president of Storage and Hyperconverged Infrastructure, said in a statement:  "Data portability is a key imperative for organizations building and deploying cloud-native applications across private and multiple clouds. NooBaa\'s technologies will augment our portfolio and strengthen our ability to meet the needs of developers in today\'s hybrid and multicloud world. We are thrilled to welcome a technical team of nine to the Red Hat family as we work together to further solidify Red Hat as a leading provider of open hybrid-cloud technologies."  Related stories:  Kidderminster-based Renault UK Clio Cup ace Dan Rowbottom will join Ciceley Motorsport for the 2019 British Touring Car Championship.  Backed by Cataclean, the \textcolor{red}{lead valuable additive to clean and fuel engine restore} and exhaust systems, Rowbottom will graduate from the Renault UK Clio Cup into one of Ciceley’s Mercedes-Benz A-Class cars for the forthcoming campaign.  He was a triple race winner last season his way to fourth place} & {The Access nodes and storage daemons make up a data plane, while the core provides its control plane.  Also: How IBM Watson is revolutionizing 10 industries TechRepublic  So, what does all mean for customers? It\'s multi-cloud storage management, which enables allows you to manage, deploy, and migrate data storage across private and major public clouds. This includes Alibaba, AWS, Azure, and Google Cloud.  It\'s easy to see why Red Hat values this. It gives their customers a way to manage storage without sweating the details across multiple platforms.  As Ranga Rangachari, Red Hat\'s vice president of Storage and Hyperconverged Infrastructure, said in a statement:  "Data portability is a key imperative for organizations building and deploying cloud-native applications across private and multiple clouds. NooBaa\'s technologies will augment our portfolio and strengthen our ability to meet the needs of developers in today\'s hybrid and multicloud world. We are thrilled to welcome a technical team of nine to the Red Hat family as we work together to further solidify Red Hat as a leading provider of open hybrid-cloud technologies."  Related stories:  Kidderminster-based Renault UK Clio Cup ace Dan Rowbottom will join Ciceley Motorsport for the 2019 British Touring Car Championship.  Backed by Cataclean, the \textcolor{red}{leading fuel additive to clean and restore engine fuel} and exhaust systems, Rowbottom will graduate from the Renault UK Clio Cup into one of Ciceley’s Mercedes-Benz A-Class cars for the forthcoming campaign.  He was a triple race winner last season his way to fourth place} 
	\end{tabular}
	\caption{Effect of random prompt text reconstruction}
	\label{table_co}
\end{table*}

\begin{table*}[!htbp]
	\centering
	
	\begin{tabular}{lccccc}
		\hline
		\textbf{Dataset} & \textbf{Task} & \textbf{Source} & \textbf{Avg len} & \textbf{Metric} &  \textbf{\#data} \\
		\hline
		Qasper & Single-Document QA & Science & 4,620 & F1  & 224 \\
		MultiFieldQA & Single-Document QA & Multi-field & 4,558 & F1  & 150 \\
		HotpotQA & Multi-Doc QA & Wikipedia & 6,657 & F1  & 300 \\
		2WikiMultihopQA & Multi-Doc QA & Wikipedia & 6,146 & F1  & 300 \\

		\hline
	\end{tabular}
	\caption{LongBench-E Information}
	\label{tab:dataset-info}
\end{table*}

\section{Fine-Tuning Datasets and Model-Generated Cases}	
\label{ap4}

Table \ref{tab:dataset-info} displays information such as the sources, average lengths, and computational metrics for various tasks. Below is a sample data entry for document question answering, primarily consisting of three parts: '\textit{input}', '\textit{context}', and '\textit{answers}'. The '\textit{input}' represents the prompt or instruction, the '\textit{context}' is the surrounding text the model needs to search through, which is often lengthy, and the '\textit{answers}' represent the possible answers derived from the context. Example:

\texttt{input: "Which park is further south within Spain, Picos de Europa National Park or Timanfaya National Park?"}

\texttt{context: 'Passage 1:Lake Ercina Lake Ercina is a small highland lake … The population is 47 (INE 2016).'}

\texttt{answers: ['Timanfaya National Park']}

When using RCC-Ins-Reconstruction for instruction reconstruction inference, we concatenate the '\textit{context}' and '\textit{input}' parts of the sample with a newline character and input them into the model's encoder for compression. Simultaneously, the decoder's input is a fixed prompt:

\texttt{prompt: "system: You are a helpful assistant. user: "}

The decoder, starting with this prompt, first reconstructs the instruction and then answers the question based on it. The content generated by the model is shown in blue font:

\texttt{"system: You are a helpful assistant. user: \textcolor{blue}{Which park is further south within Spain, Picos de Europa National Park or Timanfaya National Park? assistant: Timanfaya National Park"}}

The model accurately reconstructed the instruction and provided the correct answer, '\textit{Timanfaya National Park}'.

Additionally, we tested the RCC-Ins-compress model. The input to the RCC-Ins-compress encoder is identical to that of the RCC-Ins-Reconstruction, but the decoder's prompt is:

\texttt{Prompt: "Response of system:"}

Since RCC-Ins-compress has not been trained on instruction reconstruction tasks, it does not reconstruct the instruction in its output. Instead, it directly answers the question based on the mixed compressed context and instruction, which may result in the model failing to follow the instruction.The content generated by the model is shown in blue font: 

\texttt{"Response of system: \textcolor{blue}{Panic of 1797}"}

It can be seen that the model made an error in following the instructions.

\end{document}